%% file: learnAlg.tex
\documentclass[a4paper,12pt,numbers,sort&compress]{article}

\pdfoutput=1

\usepackage[paper=letterpaper,margin=1in]{geometry}

\usepackage{amsmath,amssymb,amsfonts,epsfig,cite,setspace,longtable,array,url,color}
\usepackage{todonotes}

\usepackage{hyperref,caption,setspace,multirow}

\usepackage{pdfpages,lipsum}

\begin{document}



\vskip 0.25in

\newcommand{\sref}[1]{\S~\ref{#1}}
\newcommand{\nn}{\nonumber}
\newcommand{\tr}{\mathop{\rm Tr}}
\newcommand{\ord}{\mathop{\rm Ord}}
\newcommand{\slog}{\mathop{\rm sLog}}
\newcommand{\sgn}{\mathop{\rm sgn}}

\newcommand{\comment}[1]{}

\newcommand{\cM}{{\cal M}}
\newcommand{\cW}{{\cal W}}
\newcommand{\cN}{{\cal N}}
\newcommand{\cH}{{\cal H}}
\newcommand{\cK}{{\cal K}}
\newcommand{\cY}{{\cal Y}}
\newcommand{\cZ}{{\cal Z}}
\newcommand{\cO}{{\cal O}}
\newcommand{\cA}{{\cal A}}
\newcommand{\cB}{{\cal B}}
\newcommand{\cC}{{\cal C}}
\newcommand{\cD}{{\cal D}}
\newcommand{\cE}{{\cal E}}
\newcommand{\cF}{{\cal F}}
\newcommand{\cL}{{\cal L}}
\newcommand{\cX}{{\cal X}}
\newcommand{\cV}{{\cal V}}
\newcommand{\cT}{{\cal T}}
\newcommand{\IA}{\mathbb{A}}
\newcommand{\IP}{\mathbb{P}}
\newcommand{\IQ}{\mathbb{Q}}
\newcommand{\IH}{\mathbb{H}}
\newcommand{\IR}{\mathbb{R}}
\newcommand{\IC}{\mathbb{C}}
\newcommand{\IF}{\mathbb{F}}
\newcommand{\IV}{\mathbb{V}}
\newcommand{\II}{\mathbb{I}}
\newcommand{\IZ}{\mathbb{Z}}
\newcommand{\re}{{\rm~Re}}
\newcommand{\im}{{\rm~Im}}

\newcommand{\tmat}[1]{{\tiny \left(\begin{matrix} #1 \end{matrix}\right)}}
\newcommand{\mat}[1]{\left(\begin{matrix} #1 \end{matrix}\right)}

\let\oldthebibliography=\thebibliography
\let\endoldthebibliography=\endthebibliography
\renewenvironment{thebibliography}[1]{%
\begin{oldthebibliography}{#1}%
\setlength{\parskip}{0ex}%
\setlength{\itemsep}{0ex}%
}%
{%
\end{oldthebibliography}%
}

\newtheorem{theorem}{\bf THEOREM}
\def\thetheorem{\thesection.\arabic{theorem}}
\newtheorem{definition}{\bf DEFINITION}
\def\thetheorem{\thesection.\arabic{definition}}
\newtheorem{proposition}{\bf PROPOSITION}
\def\thetheorem{\thesection.\arabic{proposition}}
\newtheorem{observation}{\bf OBSERVATION}
\def\thetheorem{\thesection.\arabic{observation}}
\newtheorem{conjecture}{\bf CONJECTURE}
\def\thetheorem{\thesection.\arabic{conjecture}}
\newtheorem{algorithm}{\bf ALGORITHM}
\def\thetheorem{\thesection.\arabic{algorithm}}

\def\theequation{\thesection.\arabic{equation}}
\newcommand{\setall}{\setcounter{equation}{0}
        \setcounter{theorem}{0}}
\newcommand{\setequation}{\setcounter{equation}{0}}
\renewcommand{\thefootnote}{\fnsymbol{footnote}}

~\\
\vskip 1cm

\begin{center}
{\Large \bf  Learning Algebraic Structures:} \\
{\large \bf Preliminary Investigations}

\medskip
\vspace{4mm}
\centerline{
{\large Yang-Hui He}$^{1,2,3}$ \&
{\large Minhyong Kim}$^{1,4,5}$
}
\vspace*{3.0ex}

\vspace{.2in}
\renewcommand{\arraystretch}{0.5} 
\begin{center}
{\small
{\it
\begin{tabular}{rl}
  ${}^{1}$ &
  Merton College, University of Oxford, OX14JD, UK\\
  ${}^{2}$ &
  Department of Mathematics, City, University of London, EC1V 0HB, UK\\
  ${}^{3}$ &
  School of Physics, Nan-Kai University, Tianjin, 300071, P.R.~China\\
  ${}^{4}$ &
  Mathematical Institute, University of Oxford, OX2 6GG, UK\\
   ${}^{5}$ &
   KIAS, 85 Hoegiro, Dongdaemun-gu, Seoul 02455, Korea
\end{tabular}
}
~\\
~\\
hey@maths.ox.ac.uk, \quad Minhyong.Kim@maths.ox.ac.uk
}\end{center}
\vspace{0.1in}

\begin{abstract}
We employ techniques of machine-learning, exemplified by support vector machines and neural classifiers, to initiate the study of whether AI can ``learn'' algebraic structures.
Using finite groups and finite rings as a concrete playground, we find that questions such as identification of simple groups by ``looking'' at the Cayley table or correctly matching addition and multiplication tables for finite rings can, at least for structures of small size, be performed by the AI, even after having been trained only on small number of cases.
These results are in tandem with recent investigations on whether AI can solve certain classes of problems in algebraic geometry.  
\end{abstract}

\end{center}

\newpage

\tableofcontents

\section{Introduction and Summary}\setall
Artificial intelligence (AI) and, especially, Machine Learning (ML) have become a vital component of scientific enquiry across the disciplines over the last decade, ranging from the discovery of the Higgs boson at CERN, to the mapping of the human genome, to the social science and patient diagnosis at the NHS, to AlphaGo's prowess in defeating humans in the game Go, etc. (cf.~\cite{aml,mitchell,enc,sch}).
While the basic techniques and concepts have been around since the early days of computing \cite{mp}, the modern revolution by using graphics processing units (GPUs) in the so-called AlexNet was a game-changer and brought the large-scale computations in deep neural networks available to the average laptop \cite{ksh}.

In 2017, four independent groups introduced this exciting field of machine-learning and data science to the long-standing problem of studying the plethora of vacua in superstring theory \cite{He:2017aed,ks,rue,chkn}.
In particular, a proposal was made in \cite{He:2017aed} to see whether AI, such as neural networks, can help with certain expensive computations in pure mathematics, especially in computational algebraic geometry.
Indeed, {\it can AI machine-learn calculations in algebraic geometry without understanding any mathematics but by pure pattern recognition}?
 
The inspiration came from the so-called ``string landscape'', where dimensional reduction of superstring theory in 10 space-time dimensions down to our 4-dimensional universe allows for a multitude of geometrical possibilities.
One standard solution is to take Calabi-Yau (CY) three-folds -- possibly endowed with extra structure such as stable holomorphic vector bundles -- and compute their cohomology groups, which correspond to explicit physical quantities in the standard model of particle physics (cf.~a pedagogical introduction in \cite{HeCY}).

As is well known, while a CY one-fold is just the torus $T^2$, a CY two-folds, the 4-torus $T^4$ and the K3 surface, CY three-folds are so far unclassified and billions have been constructed explicitly (the most impressive database being the hypersurfaces in Fano toric 4-folds \cite{KSCY}).
A major task over the years in the theoretical physics in conjunction with the algebraic geometry community has been to set up databases of the geometries and to compute the relevant topological quantities.

Unfortunately, computing cohomology groups for arbitrary algebraic varieties is computationally expensive: {\it in principle}, one could  embed into some appropriate projective space, set up the Gr\"obner basis (a double-exponential running time algorithm) for the defining ideal of the variety, use some version of the Euler short exact sequence to induce a long exact sequence in cohomology and compute all relevant maps in the sequence.
Quickly, one would find the actual computation becomes prohibitive, even with software dedicated to such problems such as Macaulay2 \cite{m2}, Singular \cite{singular}, or Sagemath \cite{sage}, in both running time and in memory usage.

Thus, the question was asked in \cite{He:2017aed} that, since at the most basic level, much of computational algebraic geometry reduces to finding (co-)kernels of integer matrices written in some basis of monomials, whether AI can spot some unseen pattern by machine-learning available, correct, results, and whereby ``guess'' at a right answer for an unseen situation. After all, tensor manipulation is the speciality of machine learning.
Surprisingly, it was found that many problems ranging from computing Hodge numbers or dimension of quiver representations can be machine-learnt in order to predict to an accuracy in the 90s percentage-wise in a matter of minutes, as opposed to a detailed computation which is beyond any practical means, even with the best available super-computers.

At the heart of it all, we are confirming the fact that there are underlying patterns in algebraic geometry, which might be amenable to yet unforeseen techniques.
It was reassuring that when similar experiments were tried to number theory, such as the simple task of predicting the next prime number, the neural network failed miserably \cite{He:2017aed}.
Meanwhile, solving the Rubik's cube, at the heart of which is the recognition of properties of the permutation group, has been met with success \cite{rubik}.

The purpose of this present paper is to see whether there are other natural classes of problems in mathematics which respond well to machine learning.
One would imagine from the outset that where matrix/tensor manipulations proliferate would be a good starting point.
To this end, representation theory of groups, or indeed, algebraic structures in general, should come to mind.
Therefore, it is expedient that we explore, in the spirit of experimental mathematics, a host of problems in this direction.

The organization of our preliminary investigations is as follows.
We begin with a brief introduction to machine-learning, setting out the techniques which we will use.
Then, we proceed to the studying of finite groups by focusing on the concrete problems
\begin{enumerate}
\item Can AI recognize that a given Latin square is the Cayley multiplication table of a finite group? 
\item Can AI recognize whether a finite group is simple just by ``looking'' at a Cayley table? 
\item Can AI guess the number of (isomorphism classes of) subgroups to a finite groups by ``looking'' at the Cayley table, and 
\item How good is AI at recognizing isomorphisms between finite groups.
\end{enumerate}
Next, we study finite rings.
Again, we test AI with the concrete problems of 
\begin{enumerate}
\item Checking isomorphism between finite rings, and 
\item Matching the multiplication table with the addition table for a given finite ring.
\end{enumerate}
Finally, we finish with conclusions and prospects.

\section*{Acknowledgments}
We thank Peter Braam and Mike Douglas for enlightening discussions as well as the organizers of the conference {\it Physics $\cap$ ML} at the Microsoft research HQ, Seattle in April 2019 for providing the unique opportunity for many friendly inter-disciplinary chats.
YHH is indebted to the Science and Technology Facilities Council, UK, for grant ST/J00037X/1.
MHL thanks EPSRC, UK for grant EP/M024830/1.

\section{Elements from Machine Learning}\setall
While we are certainly not giving a review of ML here, it is expedient to present some of the rudiments of the techniques which we will use throughout.
What we will use is referred to as {\it supervised learning}.
First, we have 
\begin{definition}
Labelled data $\cD = \{ x_i \to y_i \}$ for $i=1, 2, \ldots, N$ is a set of size $N$ (the data-size) consisting of ordered pairs: the input $x_i$ and the output $y_i$.
\end{definition}
In all our examples, $x_i$ will be specific integer matrices and $y_i$ will be {\it categorical} in the sense of being an element of some finite set.
Typically, $y_i$ will just be 0 or 1, and we have a so-called {\it binary query} or binary classification.

A standard strategy in ML to see how well an AI is performing is called {\bf cross-validation}, where
\begin{definition}
$\cD$ is split into the disjoint union of $\cT$, the training data and $\cV$, the validation data: $\cD = \cT \sqcup \cV$.
\end{definition}

An ML algorithm can be roughly thought of as a (highly non-analytic) function $f_{\{p_j\}}$ consisting of a multitude of parametres $p_j$ such that we minimize a cost function, typically the standard error
\begin{equation}
SD = \sum\limits_{\cT} (y_i - f(x_i))^2 \ ,
\end{equation}
over the parametres $p_j$, so as to attain a ``best-fit'' over the training set.
In other words, we are performing a complicated non-linear regression over the training data.
Having fixed the parametres and whereby ``trained'' the AI, we can validate by comparing the {\it predicted} results with the {\it actual} results in the validation dataset $\cV$.
Because the AI has seen only $\cT$ and $\cV$ is entirely unseen, the performance on validation is a good measure of how well the ML has truly ``learnt'' any underlying patterns, rather than ``over-fitting''.

One naive way to define {\bf precision} as a measure of goodness of fit is to simply find the \% agreement in the vector $f(x_i)$ of predicted values versus the vector of correct values $y_i$ within $\cV$; however, this does not distinguish between false positives and true negatives in the case of binary queries.
In the case of binary queries, the naive precision is also not a good measure when there is a disparity between the number of 1s and 0s.

One standard method is to compute the so-called {\bf Matthews Correlation}, or {\bf $\phi$-coefficient}, which generally considered to be a balanced measure in binary classification problems such as the one at hand \cite{matt,enc}.
The definition is as follows 
\footnote{
Another often-used measure is the
{\bf S{\o}rensen-Dice coefficient} \cite{enc}, usually denoted as $F_1$ and also called the F1-score.
We define the true positive rate (TPR) and the positive predictive rate (PPV), or the precision, as
$TPR := \frac{TP}{TP + FN}$ and $PPV := \frac{TP}{TP+FP}$.
The F1-score is given as the harmonic mean of the TPR and the PPV:
$F_1 := 2 \frac{PPV \cdot TPR}{PPV + TPR} = \frac{2 TP}{2TP + FP + FN} \in [0,1]$.
However, this score does not take false negatives into consideration.
}.
First, we establish the {\bf confusion matrix} consisting of the number of ``true positives'' (TP), the ``false positive'' (FP), the ``false negatives'' (FN) and ``true negatives'' (TN):
\begin{equation}
\begin{array}{|c|c|c|c|}\hline
	&& \multicolumn{2}{c|}{\mbox{{\scriptsize Actual}}} \\ \hline
 	&& 1 & 0 \\ \hline
 	\parbox[t]{3mm}{\multirow{3}{*}{\rotatebox[origin=c]{90}{{\scriptsize \hspace{9mm} Predicted}}}} & 1 & TP & FP \\
 	& 0 & FN & TN \\ \hline
\end{array}
\qquad
n = TP + FP  + FN + TN \ . \\
\end{equation}
Then, we have
\begin{definition}\label{phi}
The Matthews correlation is defined, in terms of the chi-squared of the confusion matrix as a contingency table, and explicitly, as
\begin{equation*}
\phi := \sqrt{\frac{\chi^2}{n}} =
\frac{TP \cdot TN - FP \cdot FN}{\sqrt{(TP+FP)(TP+FN)(TN+FP)(TN+FN)}} \in [-1,1] \ .
\end{equation*}
Here $\phi=1$ means perfect correlation, $\phi=0$ means randomly guessing and $\phi=-1$ means complete disagreement.
\end{definition}

There is a multi-category version of the Matthews coefficient.
Suppose we had not a binary classification problem wherein the data is labelled as 0 or 1, but as $0, \ldots, K$ with $K+1$ categories, then the confusion matrix $C_{ij}$ is a $(K+1)^2$ matrix and generalization of the Matthews coefficient is
\begin{equation}\label{phiGen}
\phi = \frac{\sum\limits_{k,\ell,m} C_{kk}C_{\ell m} - C_{kk}C_{\ell m}}
{
\sqrt{\sum\limits_{k} \left( \sum\limits_{\ell} C_{k \ell} \right)\left( \sum\limits_{k' : k' \neq k} \sum\limits_{\ell'} C_{k' \ell'} \right)}
\sqrt{\sum\limits_{k} \left( \sum\limits_{\ell} C_{\ell k} \right)\left( \sum\limits_{k' : k' \neq k} \sum\limits_{\ell'} C_{\ell' k'} \right)}
}
\end{equation}
where all sums are over the range $0, \ldots, K$.

With these measures of accuracy, we can establish a so-called training curve as follows
\begin{definition}
Let $\{x_i \to y_i\}_{i=1,\ldots,N}$ be $N$ data-points.
We choose cross-validation by taking a percentage $\gamma N$ of the data randomly for a chosen $\gamma \in (0,1]$ as training data $\cT$, the complement $(1-\gamma)N$ data-points will be the validation data $\cV$.
The performance of the machine-learning algorithm, measured by any of the goodness of fit such as precision or $\phi$, upon training on $\cT$ and validated on $\cV$, is a function $L(\gamma)$ of $\gamma$.
The {\bf Learning curve} is the plot of $L(\gamma)$ against $\gamma$.
\end{definition}
In practice, $\gamma$ will be chosen at discrete intervals, for instance in increments of 10\% until the full dataset is attained for training.
Furthermore, we repeat each random sample $\gamma N$ a number of times, so the learning curve has error bars associated with each of its points.

The above definition suffice to introduce the rudiments of supervised machine-learning which will be henceforth used.
In terms of the actual computation and implementation of the algorithms, we mentioned the following.

\paragraph{Implementation: }
We tried an artificial neural network (NN) with fully-connected linear and soft-max layers as well as a classifier which optimally combines decision trees, random forests, nearest neighbours and support vector machines, all of which are implemented on Mathematica \cite{math} and find that the classifier performs slightly better.
In addition, we used Python's {\it sklearn} library \cite{python} for support vector machines (SVM), especially for the binary classifiers.
We will collectively refer to these algorithms as AI.
All computations are done on an ordinary laptop and each training/validation takes on the order of seconds to minutes.

W have not performed a genetic algorithm to find the best hyper-parametres or attempted to find a best architecture for the neural networks.
This is in part due to the success already attained with the likes of straight-forward SVMs, and in part to the current lack of fundamental theory of which AI is best suited for which problem in pure mathematics.
For instance, in cohomology computations, a 3-layer-perceptron consisting of a fully connected linear, followed by an element-wise sigmoid, followed by a summation layer seems to perform very well.

\section{Learning Finite Groups}
Let us being with finite group theory.
Here, since the elements can be represented by a finite set of integer matrices, they serve as ideal inputs to ML.
We begin with testing whether a random Latin square is the Cayley table of a true finite group, before moving on to more sophisticated tasks of testing simplicity and isomorphisms.

\subsection{Finite Groups and Latin Squares}
We recall that given a finite group $G$ of size $n$, it is the standard method, due to Cayley, to represent it as an $n \times n$ matrix whose $(i,j)$-th entry is the result of the group multiplication of the elements $g_i g_j$; this matrix is the {\bf Cayley Table} (or multiplication table) $\cC$ and completely specifies $G$.
Indeed, iff $G$ is Abelian, $\cC(G)$ is symmetric.
The definition of $\cC$ is, of course, up to the ordering of the elements, though it is customary to place the identity element as the first row/column entry.
We shall adhere to the ATLAS \cite{atlas} notation, as encoded in GAP \cite{gap}, wherein the elements of $G$ are simply denoted as $\{1,2,3,\ldots,n\}$.
Thus, the first few Cayley tables are 
\begin{equation}
\left(
\begin{array}{c}
 1 \\
\end{array}
\right),\left(
\begin{array}{cc}
 1 & 2 \\
 2 & 1 \\
\end{array}
\right),\left(
\begin{array}{ccc}
 1 & 2 & 3 \\
 2 & 3 & 1 \\
 3 & 1 & 2 \\
\end{array}
\right),\left(
\begin{array}{cccc}
 1 & 2 & 3 & 4 \\
 2 & 3 & 4 & 1 \\
 3 & 4 & 1 & 2 \\
 4 & 1 & 2 & 3 \\
\end{array}
\right),
\left(
\begin{array}{cccc}
 1 & 2 & 3 & 4 \\
 2 & 1 & 4 & 3 \\
 3 & 4 & 1 & 2 \\
 4 & 3 & 2 & 1 \\
\end{array}
\right)
 \ldots \ ,
\end{equation}
corresponding to the groups $\II, C_2, C_3, C_4, D_4, \ldots$, where $C_i$ means the cyclic group of size $i$ and $D_i$, the dihedral group of size $i$.

The number of finite groups of size $n$ grows as \cite{oeisG}:
\begin{equation}
\#(G_n) = 
\{
1, 1, 1, 2, 1, 2, 1, 5, 2, 2, 1, 5, 1, 2, 1, 14,
\ldots
\} \ .
\end{equation}
Hence, the number of valid Cayley tables of size $n$, allowing for row and column permutation separately, grows as the above sequence, but multiplied by $(n!)^2$, i.e.,
\begin{equation}\label{noC}
\#(\cC_n) = 
\{
1, 2, 6, 48, 120, 1440, 5040, 201600, 725760, 7257600, 39916800, \ldots
\ldots
\}
\end{equation}

The Cayley table is a special case of a {\bf Latin Square}, $\cL$, which is an $n \times n$ matrix filled by $n$ symbols (here $1,2,\ldots,n$) each of which appears exactly once in each row and in each column (i.e., an $n \times n$ Sudoku solution).
As in the finite group case, there is no known analytic formula for the number of Latin Squares, though compared to \eqref{noC}, the growth rate is \cite{oeisL} much faster:
\begin{equation}\label{noL}
\#(\cL_n) = 
\{
1, 2, 12, 576, 161280, 812851200, 61479419904000,
\ldots
\}
\end{equation}
Therefore, {\em the probability of a Latin square being a proper Cayley table of a finite group (in any permutation) is essentially 0} starting from $n$ even as small as 5.

Sometimes, one considers a {\em reduced} (or normalized, or standardized) Latin square, whose first row and first column are both equal to the ordered set $\{1,2,3,\ldots,n\}$.
This would be similar to the standard Cayley table form as given in \cite{gap}.
The number of these reduced Latin squares will grow as $\#(\cL_n) / n! / (n-1)!$.
We will consider the generic and not necessarily reduced Latin squares because we will do all permutations of rows and columns of Cayley tables manually.

A natural question therefore arises as
\begin{quote}
{\bf Q: }
 How does one efficiently recognize a random Latin square of size $n$ to be the Cayley table of a finite group of size $n$.
\end{quote}
Indeed, associativity of group multiplication is by no means built into a Latin square.

\subsubsection{Recognizing Cayley Tables}
Let us  reconsider the problem by having the following data structure:
\begin{itemize}
\item Fix a size $n$;
\item Generate a random sample $R_L$ of Latin squares, say, of size  $|R_L| = 15,000$. 
This can be done, for instance, using a built-in package in \cite{sage};
\item Generate the Cayley tables for {\em all} groups of size $n$ (which can be done from \cite{gap}); perform $k$ random row and, independently, column permutations.
This gives us a set $R_C$ Cayley squares of size $\#(G_n) k^2$;
\item For $n \gtrsim 5$, $R_L \cap R_C = \{\}$ generically and we will label elements in the set $R_L$ with 0 and those in $R_C$ with 1.
Thus, we have a fully labelled dataset 
\[
\cD = \{L \rightarrow 0 : L \in R_L; \ C \rightarrow 1 : C \in R_C\} \ .
\]
\end{itemize}

With our data labelled by 0s and 1s, we can now proceed to supervised machine learning of a binary classification problem.
Now, the naive precision is not sufficient as a measure of goodness of fit since Cayley tables are rare in the space of Latin squares because of the relative paucity of positive hits.
Thus, we will need to calculate both accuracy and the $\phi$-coefficient as a measure.

\paragraph{Remark: }
The fact that we permute the rows/columns of the Cayley tables to get a balanced tally of 0s and 1s is a standard method in data-enhancement.
In classification problems in AI, this is referred to as Synthetic Minority Oversampling Technique (SMOTE) \cite{smote} and is widely used.

\begin{algorithm}\label{alg:cayley1}
Specifically, we use the following algorithm:
\begin{itemize}
\item Take a random subset $\cT \subset \cD$; this is our training set;
\item The complement $\cD - \cT$ is the validation set $\cV$;
\item Establish a classifier function $F$ on $\cT$ which minimizes the mean square error of $F(t)$ and the true value (0 or 1) for all $t \in \cT$;
\item Compute $\mbox{round}(F(v))$ for all $v \in V$; we need to round because $F(v)$ is usually a real number after the minimization on the parametres of $F$.
Calculate the accuracy and the Matthews coefficient $\phi$ for $\mbox{round}(F(v))$ versus the actual validation value of 0 or 1;
\item Repeat the above, say, for 10 times, and take the mean and standard deviation of the precision and the $\phi$ values.
\end{itemize}
\end{algorithm}
To get an idea of the performance as we increase training size, we will also perform Algorithm \ref{alg:cayley1} by varying the size of the random training set $\cT$ and plot the measures of accuracy (with error bars) as a function of $|\cT|$, this will be our {\bf training curve}.

Now, we need to be a little careful that the ML algorithm is not just recognizing permutations but is really distinguishing true Cayley tables.
To this end, we can make sure that the AI sees only some finite groups but not others.
\begin{algorithm}\label{alg:cayley2}
As an extra pre-caution that the classifier is not just learning permutations, we also perform
\begin{itemize}
\item In the case there is more than one group of size $n$, take only a subset thereof and consider the training set $\cT$ consisting of $\rightarrow 1$ only for permutations of the Cayley table of this subset of groups;
\item Establish the classifier function $F$ on $\cT$; thus the machine will not have seen the remaining groups of size $n$;
\item Compute the precision of the Cayley tables of the unseen groups by checking how many $F(v)$ equals 1.
Here we do not need to compute the Matthews coefficient because the validation data are all 1s since we are checking only against permutations of correct but unseen Cayley tables;
\item Again, repeat 10 times to obtain mean and standard deviation of $p$. 
\end{itemize}
\end{algorithm}

\begin{figure}[!h!t!b]
\includegraphics[trim=15mm 0mm 0mm 0mm, clip, width=6.5in]{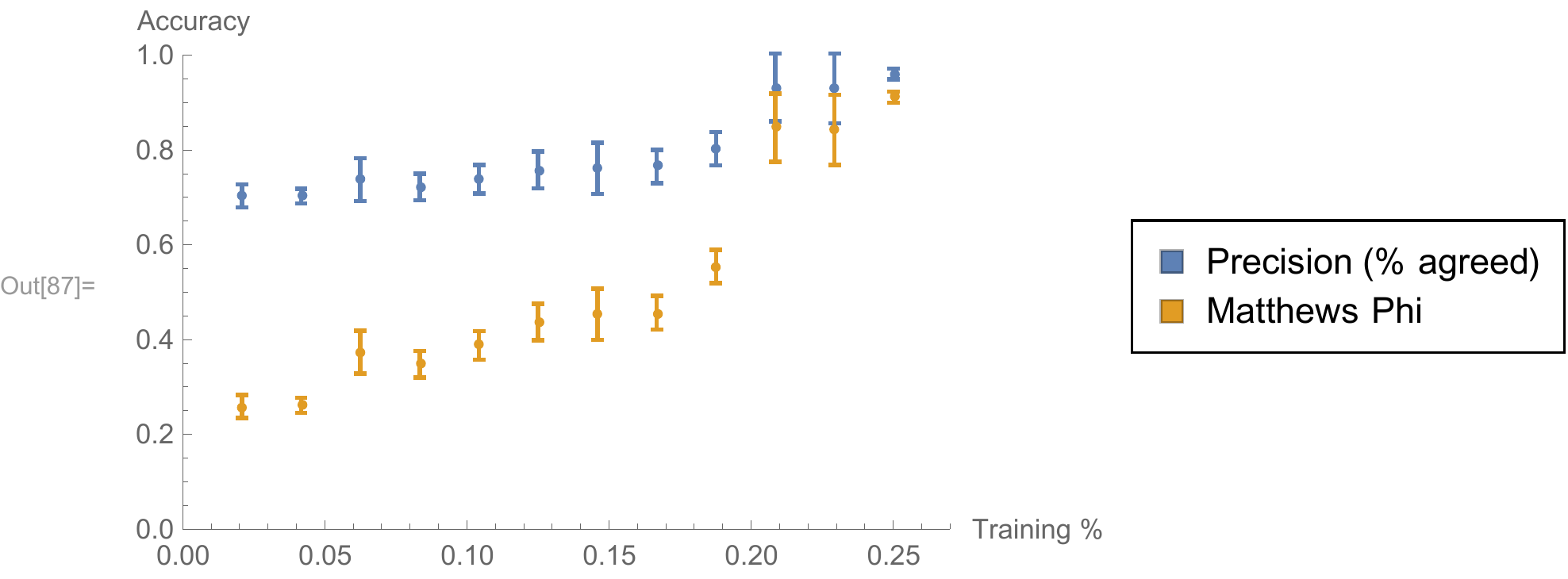}
\caption{
{\sf {\small 
The learning curve for $n=8$ Latin squares.
We training a percentage of Latin squares, some of which are valid Cayley tables. This is then validated against the remaining unseen tables.
The accuracy is measured by (1) precision and (2) the Matthews correlation coefficient $\phi$, both of which already tend to 1 at about 1/4 of the data for training.
}}
\label{f:training8}
}
\end{figure}

Thus armed, we can experiment with actual finite groups.
For concreteness, let us consider groups of size 8 and 12, of which there are a few.
\paragraph{Testing $n=8$: }
There are 5 non-isomorphic groups of order 12, viz.,
\begin{equation}\label{G8}
C_8, \ C_4 \times C_2, \ D_4, \ Q_{8}, \ C_2 \times C_2 \times C_2
\end{equation}
where $Q_8$ is the the quaternion group of size 8.

Testing Algorithm \ref{alg:cayley1} for 10 times, with $|\cT| = 5000$ (which is about 20\% of the data) and with $k=40$ random permutations on the rows and columns on the correct Cayley tables, gives, against the validation data (the remaining 80\%) $\phi = 0.914 \pm 0.027$, which very good indeed.
In other words, from a dataset consisting of 15,000 random Latin squares which are not Cayley tables as well as $40^2 \cdot 5$ which are, we took a random sample of 5000 to train the AI.
Upon checking with the remaining {\it unseen} validation set of $23,000-5,000=18,000$, constituting some 80\% of the total, we find that the binary classifier has correctly guessed at what should be a Cayley table to some 90\% confidence.

We can now vary the training size, say from 500 to 6,000 (of the total of some 20,000 data points) in increments of 500 and visualize the $\phi$-coefficient (against the unseen data) as the training curve.
This is shown in Figure \ref{f:training8}.
The error bars show standard deviation from 10 repeats for each training \%.
Both measures of the accuracy are used, the naive ``precision'', which is the \% of agreement of predicted and actual, as well as the more rigorous ``Matthews $\phi$-coefficient'' given in Definition~\ref{phi}.  The precision appears to be better though we need to be careful since there might be only a small number of true positives. In any case, both measures are shown to increase to almost 100\% as we go only up to 25\% of the total data size.

As a further check, we apply Algorithm \ref{alg:cayley2}, say, with $30^2$ permutations on the Cayley tables and with 2000 random training samples.
We now take, out of the 5 groups in \eqref{G8}, a set $S \in \{1,2,3,4,5\}$, which, combined with the random Latin squares, constitute the training data.
We then find the percentage agreement with the permutations of the Cayley tables of the remaining unseen groups.
Again, taking 10 random trials, we find that
\begin{align}
\nn
&S = \{1,2,4\} \Rightarrow p = 75.94 \pm 11.65\%
\ , \quad
S = \{1,2\} \Rightarrow p = 49.35 \pm 27.36 \%
\ , \\
&S = \{3,4,5\} \Rightarrow p = 73.93 \pm 13.58 \%
\end{align}
We see that indeed with fewer valid groups the accuracy decreases. Nevertheless, even with only 3 of the 5 groups seen and with about 10\% of total data, the remaining two groups can still be predicted with about 76\% accuracy.

\paragraph{Testing $n=12$: }
Let us try groups of a different size.
There are 5 non-isomorphic groups of order 12, viz.,
\begin{equation}
C_3 : C_4, \ C_{12}, \ A_4, \ D_{12}, \ C_6 \times C_2
\end{equation}
where $A_i$ is the alternating group on $i$ elements and $G_1:G_2$ is the  semi-direct product of the two (here, $C_3 : C_4$ gives the dicyclic group of order 12).
We generate 20,000 random Latin squares, 2500 random permutations of each of the 5 true multiplication tables.
Then, applying Algorithm \ref{alg:cayley1} with 5000 random labelled data points as training and validating against the remaining 15,000, we find that 
the results are just as good, i.e., with only a 1/4 sized training data we have
\begin{equation}
\mbox{Precision } = 94.98 \pm 0.93 \% \ ,
\qquad
\mbox{Matthews } \phi = 0.90 \pm 0.02 \ .
\end{equation}

Let us remark on an interesting phenomenon.
Let us first rethink a little about the configuration matrices themselves.
For a finite group/Latin square of size $n$, the entries of the $n \times n$ matrices are $1, 2, \ldots n$.
It is clear from the above investigations that the AI is very good at recognizing permutations with ascribed properties.
What if we adjusted the entries?
That is, suppose we changed the entries for e{\it each} permutation, by, for instance, increasing all entries by 1.
This way, we are somewhat ensuring that permutations themselves are not immediately visible.

For example, suppose we have a Cayley table for a group of size 12 (a $12 \times 12$ Latin square with entries from 1 to 12), then we perform a random permutation of its rows and columns, and at the same time add 1 to all the entries, we now have  a $12 \times 12$ Latin square with entries 2 to 13.
We repeat the random permutations, each time increasing the entries by 1.
All these cases will then be labelled as ``correct'' (or ``1'') because they come from true Cayley tables.
Similarly, we take random Cayley tables with such augmentation, i.e., we choose one particular Latin square which does {\it not} correspond to the Cayley table of a finite group, which will be a $12 \times 12$ Latin square with entries from 1 to 12, and then perform random permutation of its rows and columns, each time adding 1 to all the entries.
These will be labelled as ``incorrect'' (or ``0'').
Thereby we obtain a data list of of size around 25,000.

Again, training with 10,000 cases and validating against the remaining 15,000, we achieve 69.04\% precision with $\phi = 0.21$.
The performance is worse than before, as expected, since now there is a {\it huge} variation of in the entries: starting from 1 to 12, up to around 25,000.
It is still interesting that the classifier can do as well as near 70\% precision (albeit the confidence $\phi$ is rather low).
We do not expect that increasing data size would increase the precision since each time we perform a permutation as we augment the data, we are also increasing the entries of the matrices.

\subsubsection{Known Theorems}
What we have done here with Cayley table and Latin squares is, of course, a warm-up.
There are known results  (q.v.~Theorem 1.2.1 of \cite{kd}):
\begin{theorem}\label{quad}
A Latin square $a_{ij}$ is the Cayley multiplication table of a finite group IFF the {\em quadrangle criterion} holds, i.e., for any indices $i,j,k,l$ and $i,j,k,l$, $a_{ik} = a_{i'k'}, \ a_{il} = a_{i'l'}$  and $a_{jk} = a_{j'k'}, \ a_{jl} = a_{j'l'}$.
\end{theorem}
As one can see, this is an explicit permutation criterion.
Therefore, what our classifier is learning and recognizing, is the permutation of row/columns (something in which a linear hidden layer or a decision tree excels) as well as the above rule in shuffling indices (which supposedly the neural network is also good at).

That said, it is still quite remarkable.
The AI has no idea about the quadrangle criterion, or for that matter, anything about group theory.
It is only a pattern recognition device for matrices.
Yet, from gathering information on some 10,000 data points of when a Latin square {\em should} be a proper Cayley table, it is effectively {\it predicting} the correct criteria.

\subsection{Properties of Finite Groups}\label{s:group}
Having gained some confidence with the defining structure of finite groups via the Cayley table, it is expedient to wonder whether other important quantities of finite groups can be machine-learnt.
Here, we will test a few of these properties which are categorical in that they can be discretely classified.

\subsubsection{Simple Groups}

\begin{figure}[!h!t!b]
\includegraphics[trim=15mm 0mm 0mm 0mm, clip, width=6.5in]{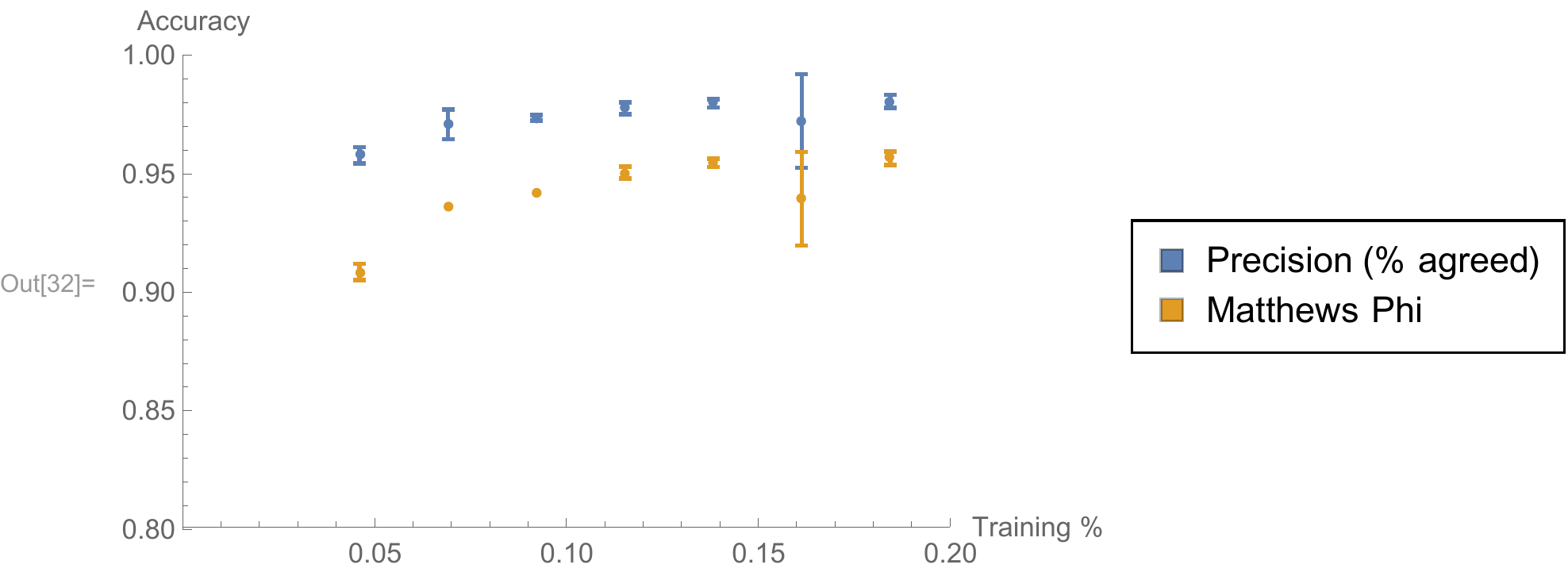}
\caption{
{\sf {\small 
The learning curve for recognizing whether a group is simple.
The data consists of (permutations, with enhanced frequencies for the simples, which are relatively rare) all Cayley tables of all first 602 finite groups, up to size 70.
We are nearing complete agreement by seeing up to only 20\% of total data.
}}
\label{f:trainingSimple}
}
\end{figure}

One of the most essential characteristic of a finite group is whether it is simple.
Can this be tested efficiently just from ``looking'' at the Cayley table?
Note that this is a highly non-trivial task: testing whether a group is simple just by knowing the multiplication table is not straight-forward.
\begin{algorithm}
As a concrete database, we take all finite groups up to and including size 70.
This is a total of 602 groups.
\begin{itemize}
\item We compute all the Cayley tables and bottom-right-pad them all into $70 \times 70$ matrices; i.e., we embed all Cayley tables into $70 \times 70$, our maximal size, corresponding to the groups of size 70, into the top-left corner and setting all remaining entries for any tables of dimension less as 0;
\item As before, we permute each (row/column independently) for, say $10^2$ times, labelling all (permutations of) those which are simple as 1 and non-simple as 0; this is a list of some 60,000 binary-labelled data points;
\item Let us take a random training sample of $\gamma \%$ of the data points;
\item We validate against the remaining $(1 - \gamma )\%$.
\end{itemize}
\end{algorithm}
For instance, taking 5000 points to train (amounting to less than 10\% seen of the full data),
we find that the AI has learnt and validated to up 98.77\% precision, with $\phi = 0.76$.
This is very impressive, for  5 - 10 minutes of computation time on an ordinary laptop.

Again, we remark that, crucially, there is a great paucity of simple groups relative to non-simples: in our 602 sample, only 20 are actually simple.
This is a situation of so-called imbalanced data.
To make sure it is not this imbalance we are accidentally picking up, we again use SMOTE \cite{smote}.

Here, since we are doing permutations to generate more sample size in any event,  we can take more row/column permutations for the Cayley tables of the under-represented simple groups.
For example, taking $20^2$ random permutations for simples and $5^2$ for non-simples gives a data size of some 20,000, about one-third of which can be labelled as 1 (for simple), no longer  a minority.
This is very much similar to a recent study of distinguishing elliptically fibred Calabi-Yau manifolds \cite{He:2019vsj}: the number of known non-elliptic-fibration is in the great minority and SMOTE is applied by permutations of configurations which lead to equivalent manifolds.

With our more balanced dataset, training on 5000 random samples (less than 1/4) and validating against the remaining 3/4 gives a precision of 98.36\% and with $\phi = 0.96$.
The learning curve of the first 5\% to 20\% training size, validated against the remaining 95\% to 80\% are shown in Figure \ref{f:trainingSimple}.
Thus now we can indeed say with confidence that the AI has ``learnt'' which Cayley tables are those of simple finite groups.

\subsubsection{Number of Subgroups}

The number of non-isomorphic subgroups $|L_G|$ for a given finite group $G$ is also a wildly fluctuating quantity
(the set of non-isomorphic subgroups is sometimes called the lattice of subgroups).
In our first 602 finite groups up to size 70, this number varies from 2 (since $\{1\}$ and the group itself are always trivial subgroups, we actually include $\{1\}$ also as a group and here is the single instance where the number of subgroups is equal to 1) to 2825.
Due to this large fluctuation, let us categorize the number to be
\begin{equation}\label{subgroup3}
\left\{
\begin{array}{lcl}
0 &:& |L_G| < 30 \ ; \\ 
1 &:& 30 \leq |L_G| < 100 \ ; \\
2 &:& 100 \leq |L_G| \ .
\end{array}
\right.
\end{equation}
This division is entirely a choice, of course, set up mainly to test how the AI responds to multi-category classification.

\begin{figure}[!h!t!b]
\includegraphics[trim=0mm 0mm 0mm 0mm, clip, width=6.5in]{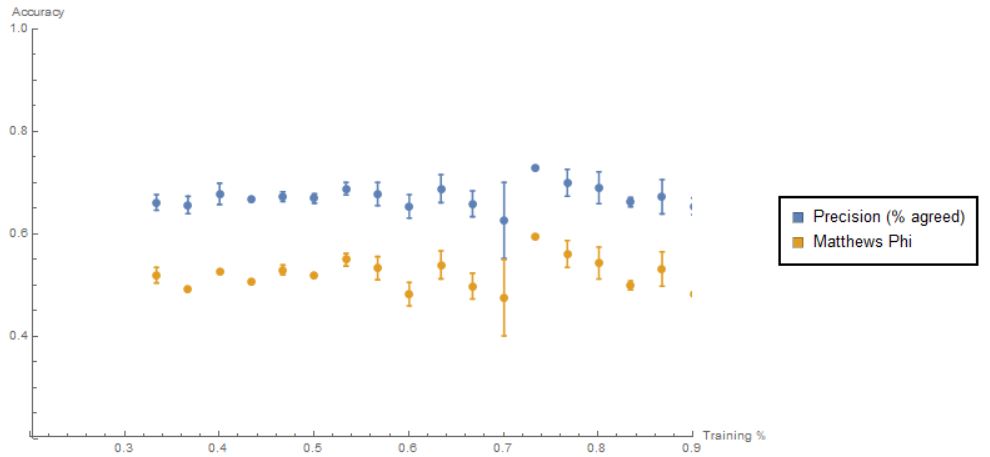}
\caption{
{\sf {\small 
The training curve for categorizing number of subgroups for all the first 602 finite groups.
}}
\label{f:trainingSubgp}
}
\end{figure}

We take each of the 602 Cayley table and perform $5^2$ random row/column permutations, labelled with $0,1,2$ according to how many subgroups there are for the group using \eqref{subgroup3}, and then completely shuffle the order of the $602 \cdot 5^2$ data-points.
This constitutes our labelled dataset $\cD$.
As always, we split $\cD$ into training and validation sets and plot the training curve for both the precision and the phi-coefficient.
This is shown in Figure \ref{f:trainingSubgp}.
The performance is not stellar, with accuracy hovering at the 60 -  80\% range.
Nevertheless, given the difficulty of the problem, it is interesting that the AI has learnt something at all.

\subsubsection{Isomorphisms}\label{s:groupiso}
Next, we test whether the AI can recognize isomorphism between finite groups.
Let us test with the cohort of the 5 groups of size 12 and apply the following
\begin{algorithm}
Split the 5 groups into two sets, say $S_1 = \{ C_3 : C_4, \ C_{12}, \ A_4 \}$ and $S_2 = \{ D_{12}, \ C_6 \times C_2 \}$.
\begin{itemize}
\item Generate random row/column permutations of the Cayley tables of the groups in $S_1$, again increasing all entries by 1 for each permutation.
	Consider pairs of such matrices, labelling as 1 if they correspond to the same group and 0, if not.
	This constitutes the training data for isomorphism;
\item Do the same for the groups in $S_2$, this is the validation data.
\item For concreteness we take equal size training and validation, of, say, 20,000.
\item Compute the precision and phi-coefficient, with errors gathered from 10 trials.
\end{itemize}
\end{algorithm}

Applying the above calculation, we find that, unfortunately and rather interestingly, that AI achieves only $49.12 \pm 5.7 \%$ precision, with $\phi = 
0.0003 \pm 0.008$, this is clearly not any better than random guessing.
Therefore, it seems that the isomorphism problem is {\it not} learnable, unlike some of the other problems we have encountered.

\section{Finite Rings}\setall
Having gained experience with finite groups, it is expedient to try more complicated algebraic structures.
Let us consider finite rings $R$.
Once again, finite rings of small size are classified \cite{oeisR}, the growth rate of the number of inequivalent rings of a given size $n$ proceeds as 
\begin{equation}
\#(R_n) = 1, 2, 2, 11, 2, 4, 2, 52, 11, 4, 2, 22, 2, 4, 4, 390, \ldots 
\end{equation}
Being a ring, we are now endowed with both multiplication ``*'' and addition ``+''.
Therefore the analogue Cayley table is a pair of tables, one for ``*'' and one for ``+''.
Since the additive identity ``0'' multiplies any element to ``0'', there is no reason why the multiplication table should be a Latin square. 
Hence, we will adopt a different strategy, which is still along the same vein.

\subsection{Non-Isomorphic Rings of a Given Order}
To be concrete and computationally feasible let us focus on commutative rings coming from direct products of cyclic groups, i.e.,
\begin{equation}\label{ringR}
R = \IZ/n_1\IZ \times \IZ/n_2\IZ \times \ldots \times \IZ/n_k\IZ \ ,
\end{equation}
wherein multiplication and addition are the usual arithmetic operation, but modulo $(n_1, n_2, \ldots, n_k)$ respectively.
For example, the ring $R = \IZ/2\IZ \times \IZ/3\IZ$ has six elements that can be written as
\begin{equation}
\{ (0,0),\ (0,1),\ (0,2),\ (1,0),\ (1,1),\ (1,2) \}
\end{equation}
modulo 2 and 3 respectively, as a 2-vector.
We can label these elements as 1 to 6.
Multiplication and addition thus proceeds readily as
\begin{equation}
* : \
\left(
\begin{array}{cccccc}
 1 & 1 & 1 & 1 & 1 & 1 \\
 1 & 2 & 3 & 1 & 2 & 3 \\
 1 & 3 & 2 & 1 & 3 & 2 \\
 1 & 1 & 1 & 4 & 4 & 4 \\
 1 & 2 & 3 & 4 & 5 & 6 \\
 1 & 3 & 2 & 4 & 6 & 5 \\
\end{array}
\right)
\ ;
\qquad
+ : \
\left(
\begin{array}{cccccc}
 1 & 2 & 3 & 4 & 5 & 6 \\
 2 & 3 & 1 & 5 & 6 & 4 \\
 3 & 1 & 2 & 6 & 4 & 5 \\
 4 & 5 & 6 & 1 & 2 & 3 \\
 5 & 6 & 4 & 2 & 3 & 1 \\
 6 & 4 & 5 & 3 & 1 & 2 \\
\end{array}
\right)
\end{equation}
We see that the ``+'' table is a Cayley table for the Abelian group $C_6$ but the ``*'' table is not a Latin square at all.

Now, in order to generate a large number of non-isomorphic rings, let us fix a prime, say $p=2$ and consider rings of the form
\begin{equation}\label{ring2n}
R_n = \prod\limits_{i=1}^k \IZ / 2^{n_i} \IZ \ , \qquad N = \sum\limits_{i=1}^k n_i \ .
\end{equation}
This is a finite ring of size $n = 2^N$.
In particular, every integer partition of $N$ gives a truly different $R$.

We can now proceed to see whether the AI can tell whether a pair of multiplication and addition tables correctly reflect the structure of a particular finite ring.
\begin{algorithm}\label{alg:ring2^n}
To test whether multiplication and addition tables are correlated correctly,
\begin{itemize}
\item We generate all possible partitions of $N$, giving us $\pi(N)$ non-isomorphic rings of size $n=2^N$, for each, we compute the pair of multiplication table and the addition table, $(M,A)$, both of size $n \times n$;
\item Consider $K$ random permutations of the row and column of $M$ and $A$ where the rows and columns are independently permuted, though the row and column permutations are the same for $M$ and $A$ respectively;
\item We will denote the set of pairs $\{ (M_j, A_j) \}_{j=1,2,\ldots,K}$ as $Co$ for ``correct''; these correspond to correct multiplication/addition tables of our finite rings;
\item Now generate {\it mismatchings} by having take random combinations of $M_j$ and $A_{j'}$ from Co for $j \neq j'$. We will denote the set of pairs 
$\{ M_j, A_{j'} ) \}$ as $Inc$ for ``incorrect''; these correspond to nonsensical pairs and do not correspond to the multiplication/addition tables of any finite ring;
\item We now establish the dataset $\cD =  Co \sqcup Inc$ where elements in $Co$ are labelled as 1 and those of $Inc$, as 0.  
\item We can now split $\cD$ randomly as a training set $T$ and validation set $D$ with $D$ being the disjoint union $\cD = \cT \sqcup \cV$.
\item Train the AI (an optimized combination of decision trees, random forests and support vector machines, etc.) on $\cT$ and validate against $\cV$, check the precision as well as the Matthews $\phi$-coefficient.
\item Repeat the above for, say 10 times, to get a mean and SD.
\end{itemize}
\end{algorithm}

Testing Algorithm \ref{alg:ring2^n} with $n=6$, training size of 5000 and validation size 14400-5000, we find the precision to be $69.08 \pm 0.33 \%$, though the Matthews coefficient is only $\phi = 0.037 \pm 0.017$.
Again, this is not better than random guessing.
Thus it appears that like checking isomorphisms between finite groups, a ring structure is not immediately recognizable by our implementation of the AI.

\subsection{Learning a Single Ring}
Let us return to \eqref{ringR}, but in a slightly different context to test machine learning.

\begin{algorithm}\label{alg:singleRing}
Focusing on a single given finite ring,
\begin{itemize}
\item Fix $n$ and consider a factorization thereof so that $n = \prod\limits_{i=1}^k n_i$ where each $n_i$ is prime, possibly with multiplicity;
\item Generate the pair $(M,A)$, the $n \times n$ multiplication and addition table of the finite ring \eqref{ringR};
\item Consider $K$ random permutations of the row and column of $M$ and $A$ where the rows and columns are independently permuted, though the row and column permutations are the same for $M$ and $A$ respectively; these permutations should be the same as randomly permuting the ordering $n_1, n_2, \ldots, n_k$, and hence different orders of the prime factorization of $n$.
We will denote the set of pairs $\{ (M_j, A_j) \}_{j=1,2,\ldots,K}$ as $Co$ for ``correct''; these correspond to correct multiplication/addition tables of our finite rings;
\item Now generate {\it mismatchings} by having take random combinations of $M_j$ and $A_{j'}$ from Co for $j \neq j'$. We will denote the set of pairs 
$\{ M_j, A_{j'} ) \}$ as $Inc$ for ``incorrect''; these correspond to nonsensical pairs and do not correspond to the multiplication/addition tables of any finite ring;
\item We now establish the dataset $\cD =  Co \sqcup Inc$ where elements in $Co$ are labelled as 1 and those of $Inc$, as 0.  We can now split $\cD$ randomly as a training set $\cT$ and validation set $\cV$ so that  $\cD = \cT \sqcup \cV$.
\item Train the AI on $\cT$ and validate against $\cV$, check the precision as well as the Matthews $\phi$-coefficient.
\item Repeat the above for, say 10 times, to get a mean and SD.
\end{itemize}
\end{algorithm}

\begin{figure}[!h!t!b]
\includegraphics[trim=15mm 0mm 0mm 0mm, clip, width=6.5in]{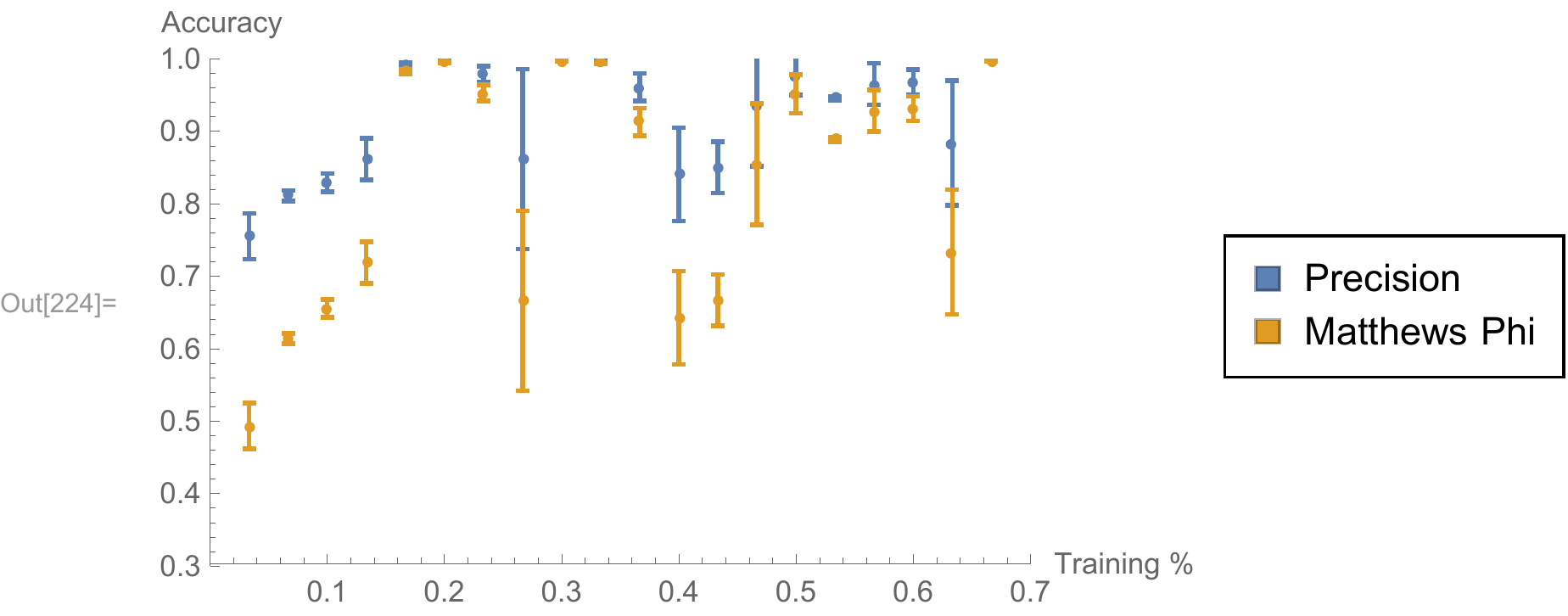}
\caption{
{\sf {\small 
The training curve for $N=12$, corresponding to the finite ring $R=\IZ/2\IZ \times \IZ/2\IZ \times \IZ/3\IZ$.
In steps of 1000, we proceed from 1000 to 20,000 random sample out of a total of 30,000 as training data.
This is validated against the remaining complementary set.
}}
\label{f:trainingR12}
}
\end{figure}

\paragraph{Checking $n=12$}
For concreteness, let us consider a finite ring of size $12$, viz., $R = \IZ/2\IZ \times \IZ/2\IZ \times \IZ/3\IZ$.
We generate 2,500 random permutations for the set $Co$ and 10,000 random permutations for the set $Inc$, giving a dataset of size 12,500.
Training 4,000 samples (about 1/3) and validating against the remaing 12,500 - 4,000 = 8,500, we find the precision to be $87.6 \pm 0.5 \%$ and the Matthews $\phi = 0.56 \pm 0.02$, not bad for having only seen 1/3 of the data.

As before, we can see the development of accuracy versus training sample size in the training curve, which is presented in Figure \ref{f:trainingR12}.
In steps of 1000, we proceed from 1000 to 20,000 random sample out of a total of 30,000 as training data.
This is validated against the remaining complementary set.
Both precision (\% agreement) and Matthews $\phi$-coefficient are shown for each case, together with error bars coming from the standard deviation of over 10 random repeats.
While the performance does not seem as good as the Cayley table case, it is still very impressive, with both precision and $\phi$ in the 80 - 95 \% range at less than 50\% of total data.
Therefore, it seems that once a particular ring is fixed, random associations between addition and multiplication tables can be distinguished between true pairings.
But perhaps the AI is merely distinguishing permutations. To ensure that this is not so, we mix up a collection of rings and validate on rings it has {\it not} seen before.  We do so in the next subsection.

\subsubsection{A Collection of Rings}
We now try to see whether the AI can indeed recognize ring structure by testing it with rings it has never encountered before as follows:
\begin{algorithm}
We split the rings into unseen and seen, not just the permutations, as follows:
\begin{itemize}
\item Take a fixed integer $N$ and consider all rings of the form $\prod_i \IZ / n_i \IZ$ up to $N$, i.e., we consider all prime factorization of $j=2, \ldots, N$.
This will give us $N-1$ different finite rings.
\item Split the rings into the first  $F$ and the last $N-1-F$. Establish the multiplication/addition tables, but {\it padded} so that everything is $N \times N$; i.e., we embed the tables into $N \times N$ so that everything bigger than the actual size of the tables is padded right and down;
\item The training data will be the list of correctly matched multiplication/addition tables as well as their random (correlated) permutations of row/columns, labelled with 0 and 1 as in Algorithm \ref{alg:singleRing} for {\it only the first $F$ rings}; the validation data will be the same, but for {\it only the last $N-1-F$ rings}.
\end{itemize}
\end{algorithm}

To be concrete, set $N=10$, training with the (permutations of) the first $F=7$ rings and validating against the remaining $2$.
A total size of 37,500 is established and we find that
\begin{equation}
\mbox{Precision } = 80.40 \% \ ,
\qquad
\mbox{Matthews } \phi = 0.6520 \ .
\
\end{equation}
Therefore, indeed the AI is genuinely learning about the matching of multiplication and addition tables, the very underlying definition of the structure of a finite ring.
It is interesting, however, to note that unlike Figures \ref{f:training8} and \ref{f:trainingSimple}, the learning curve in Figure \ref{f:trainingR12} fluctuates more wildly in addition to performing to less accuracy/precision.

\section{Conclusions and Prospects}
We have taken a preliminary step in the investigation of the question whether AI can be used to identify patterns in algebraic structures.
In particular, we focused on finite groups and finite rings, especially on isomorphisms, subgroup structures and recognition of properties.
The warm-up exercise of whether given a Latin square, AI can recognize it to be the Cayley table of a finite group was very assuring.
Here, using finite groups of order 8 and 12 as test-ground, we find that ML can ``learn'' that a Latin is truly a Cayley table to precision/accuracy more than 90\%.
While there is a known theorem of testing this, the standard algorithm is itself not efficient since many permutations need to be performed for the explicit check.
The AI has somehow managed to do the identification without knowing anything about groups.

More impressive is the result of the experiment of guessing whether a finite group is simple by merely ``looking'' at its Cayley table.
Indeed, the actual algorithm to do this is not at all straight-forward.
Nevertheless, within seconds and examining the cases of only about 10\% of the conglomerate of (permutations) of all the groups up to order 70, the AI has seemingly learnt the correct pattern and could predict the remaining 90\% of the unseen cases to over 98\% precision.

These phenomena are very much in the spirit of \cite{He:2017aed,Bull:2018uow} when ML was used to mysteriously compute bundle cohomology over Calabi-Yau varieties.
The standard algorithms for these involve exponentially expensive Gr\"obner basis computations while the AI could guess to over 90\% accuracy the correct ranks of the cohomology groups, whereby giving such quantities as Betti numbers and Hodge numbers.

Whilst these preliminary results are very encouraging, some other experiments show that treatment of similar problems, mutatis mutandis, does not furnish as good results.
For instance, checking whether two finite groups or two finite rings are isomorphic renders the sort of support vector machines and classifiers ineffective at Matthew's phi-coefficient essentially 0.
Also, categorizing the number of subgroups of a given finite group $G$ also seems less learnable.  This seems to be related to the fact that this number depends sensitively to number-theoretical properties of the order of $G$ such as its factorization and anything involving predicting properties of primes by AI is unsurprisingly a difficult one \cite{He:2017aed}.

Nevertheless, for a given ring, mixing up the pairs of multiplication and addition tables by random permutation, only some of which will be correctly matched, give a dataset on which the AI does indeed work to high precision again (up to 90\% precision at only a small percentage training).

It should be emphasized that the AI is doing more than merely recognizing permutations.
In fact, there is no current good ML techniques \cite{enc} which are adapted to recognizing random permutations.
What the AI is spotting are some inherent patterns in the structure, such as Theorem \ref{quad} which underlie the question of when a Latin square is a Cayley table.
Indeed, recently in light of whether AI can distinguish elliptic fibration structure over classes of Calabi-Yau manifolds \cite{He:2019vsj}, a control experiment of randomly assigning yes/no was applied and reassuringly the AI (there in the form of a neural network and a neural classifier) was effective to Matthew's phi-coefficient around 0, no better than arbitrarily guessing.
Thus indeed, when the AI does recognize a pattern, the latter is one which is of some inherent meaning.

As is almost always with machine-learning, the results are both encouraging and mysterious.
Whilst we are slowly and steadily compiling the list of problems in mathematics which can be done by AI much more efficiently than with traditional algorithms, many of which are exponentially complex, the question remains as to what exactly is the AI doing.

At the most basic level, be they neural networks or SVMs, the AI has simply found a (in principle highly non-explicit) function which best fits some inputs to outputs.
The key is to understand, even in part, such a function, because whatever it may be, it is {\it bypassing} traditional approaches.
Calculating cohomology groups of an algebraic variety typically requires many terms in long-exact sequence chasing, likewise, deciding whether a finite group is simple requires delicate usage of Sylow theorems. Yet, gathering from the available data of these brute-force attacks accumulated in the literature over the years, the AI seems to be circumventing (at least statistically) these intensive algorithms, by simply ``looking'', ``fitting'' and ``guessing'',  orders of magnitude faster.
Should we understand this function, we would arrive at new theoretical shortcuts in many explicit computations.

\input biblio

\end{document}